\documentclass[11pt]{article}

\usepackage{acl}
 \usepackage{microtype}
\usepackage{graphicx} 
\usepackage{listings}
\usepackage{tcolorbox}
\tcbuselibrary{breakable}
\usepackage{setspace}
\usepackage{float}
\usepackage{booktabs}

\usepackage[english,bidi=default]{babel} % English as the main language.
\babelfont{rm}[
  BoldFont=texgyretermes-bold.otf,
  ItalicFont=texgyretermes-italic.otf,
  BoldItalicFont=texgyretermes-bolditalic.otf
]{texgyretermes-regular.otf}
\babelprovide[import]{hindi}
\babelfont[*devanagari]{rm}{FreeSerif.otf}
\babelprovide[import]{arabic}
\babelfont[*arabic]{rm}[
  BoldFont=Amiri-Bold.ttf,
  ItalicFont=Amiri-Italic.ttf,
  BoldItalicFont=Amiri-BoldItalic.ttf
]{Amiri-Regular.ttf}

\usepackage{xcolor}
\usepackage{multirow}
\newtcolorbox{promptbox}{
  colback=gray!5,
  colframe=black!60,
  boxrule=0.5pt,
  arc=3pt,
  left=6pt,
  right=6pt,
  top=6pt,
  bottom=6pt,
  breakable
}
\usepackage{hyperref}
\hypersetup{
  pdfcreator  = {},
  pdfproducer = {}
}
\ifdefined\pdfvariable
  \pdfvariable suppressoptionalinfo \numexpr 1+2+4+8+16+32+64+128+256+512 \relax
\fi
\usepackage{mathastext}
\title{Can Linguistically Related Languages Guide LLM Translation in Low-Resource Settings?}

\author{Aishwarya Ramasethu \\
  Prediction Guard \\
  \And
  Rohin Garg$^*$ \\
  Scale AI \\
  \And
  Niyathi Allu$^*$ \\
  Independent \\
  \AND
  Harshwardhan Fartale$^*$ \\
  Independent \\
  \And
  Dun Li Chan \\
  INTI International College Penang}
\date{\today}
\begin{document}

\maketitle
\renewcommand{\thefootnote}{\fnsymbol{footnote}}
\footnotetext[0]{$^*$Equal contribution}
\renewcommand{\thefootnote}{\arabic{footnote}}
\begin{abstract}
Large Language Models (LLMs) have achieved strong performance across many downstream tasks, yet their effectiveness in extremely low-resource machine translation remains limited. Standard adaptation techniques typically rely on large-scale parallel data or extensive fine-tuning, which are infeasible for the long tail of underrepresented languages. In this work, we investigate a more constrained question: in data-scarce settings, to what extent can linguistically similar pivot languages and few-shot demonstrations provide useful guidance for on-the-fly adaptation in LLMs? We study a data-efficient experimental setup that combines linguistically related pivot languages with few-shot in-context examples, without any parameter updates, and evaluate translation behavior under controlled conditions. Our analysis shows that while pivot-based prompting can yield improvements in certain configurations, particularly in settings where the target language is less well represented in the model’s vocabulary, the gains are often modest and sensitive to few shot example construction. For closely related or better represented varieties, we observe diminishing or inconsistent gains. Our findings provide empirical guidance on how and when inference-time prompting and pivot-based examples can be used as a lightweight alternative to fine-tuning in low-resource translation settings.
\end{abstract}

\section{Introduction}

The advent of transformer-based architectures and general-purpose Large Language Models (LLMs) such as ChatGPT~\citep{openai2024gpt4}, DeepSeek-R1~\citep{deepseekai2025deepseekr1incentivizingreasoningcapability}, Mistral~\citep{jiang2023mistral7b}, and Llama~3~\citep{llama3modelcard} has led to substantial advances in machine translation over the past decade. These models exhibit strong multilingual capabilities and, for many high-resource languages, approach expert-level translation quality. However, this performance remains highly uneven across languages.

Despite the existence of over 7{,}000 languages worldwide~\citep{Eberhard2024Ethnologue}, NLP research and model development remain heavily skewed toward a small set of high-resource languages~\citep{joshi2021statefatelinguisticdiversity,Pakray_Gelbukh_Bandyopadhyay_2025}. Prior work has documented significant disparities in LLM translation performance between English and low-resource languages~\citep{choudhury2023generative}, and recent surveys show that even state-of-the-art models such as GPT-4 often fail to outperform specialized systems on languages using non-Latin scripts~\citep{info16090723}.

To address these disparities, substantial effort has gone into expanding multilingual datasets and models. Foundational work on massively multilingual representation learning, such as mBERT and XLM-R~\citep{arivazhagan2019massivelymultilingualneuralmachine,conneau2020unsupervisedcrosslingualrepresentationlearning}, enabled cross-lingual transfer across hundreds of languages. More recent initiatives, including No Language Left Behind (NLLB)~\citep{team2022NoLL} and the FLORES benchmark~\citep{goyal2022flores}, aim to scale multilingual machine translation to previously underrepresented languages, while projects such as Aya~\citep{ustun2024aya}, BLOOM~\citep{leong-etal-2022-bloom}, and Masakhane~\citep{nekoto2020participatory} emphasize broader linguistic coverage and community-driven data creation. Despite these efforts, coverage remains uneven, and many languages with low digital presence are still only partially supported in widely deployed translation systems.

Rather than focusing on resource-intensive data collection or training new language-specific models, we investigate whether the few-shot instruction-following capabilities of existing LLMs can be leveraged for extremely low-resource machine translation. We study an inference-time approach that combines linguistically related few-shot examples with a pivot language—a higher-resource language closely related to the target—to provide additional contextual grounding during generation.

Our experiments focus on two linguistically distinct yet underrepresented languages: Tunisian Arabic (\texttt{aeb})~\citep{mahdi2025llmsunderstandtunisianarabic} and Konkani (\texttt{gom})~\citep{RAJAN2020100299}. Both languages have substantial regional and cultural significance but receive limited coverage in multilingual benchmarks and are only partially supported in large pretrained translation systems such as NLLB~\citep{team2022NoLL}. This makes them representative of practical low-resource scenarios where parallel data is scarce and model support varies across dialects and scripts. We evaluate our approach using In-Context Learning (ICL) with frozen, decoder-only LLMs.

We find that incorporating linguistically and semantically related few-shot examples can improve translation behavior in certain configurations, particularly when the target language appears weakly represented in the model’s pretraining distribution. For Konkani, pivot-augmented prompting yields moderate gains in chrF++ relative to direct prompting, while for Tunisian Arabic the improvements are smaller and less consistent across models. These results suggest that the effectiveness of pivot-guided prompting depends strongly on language relatedness, representational coverage, and interactions between pivot and target varieties, rather than offering a universally reliable translation strategy.
\section{Related Work}

\subsection{In-Context Learning}

Prior work shows that multilingual LLM translation performance under few-shot in-context learning (ICL) depends strongly on prompt example quality \citep{chowdhery2022palmscalinglanguagemodeling}. However, it is also highlighted that substantial gains are observed in the high-resource language pairs. In addition \citet{agrawal2022incontextexamplesselectionmachine} confirm that even a single noisy or semantically unrelated demonstration can drastically reduce translation quality, whereas a well-formed equivalent-meaning example is often sufficient to elicit better translation quality from the pretrained LLMs.  

Further work by \citet{vilar-etal-2023-prompting} demonstrate that translation quality depends on domain quality rather than lexical similarity of the in-context examples and that the quality of translation degrades with poorly selected in-context examples. However their evaluation is limited to translations between English and a small set of relatively high-resource languages (French, German, and Chinese). The work of \citet{pmlr-v202-garcia23a} also supports that the quality of few-shot in-context examples is crucial. \citet{puduppully2023decomposedpromptingmachinetranslation} introduce DecoMT, a few-shot prompting approach that decomposes the translation process into a sequence of word-chunk translations. 

Large-scale analyses show that ICL performance in MT is driven primarily by example quality and target-side distribution rather than prompt structure or ordering \citep{zhu2024multilingualmachinetranslationlarge,chitale-etal-2024-empirical}.

\citet{zhu-etal-2024-towards-robust} investigate robustness in ICL by introducing a dual-view demonstration selection strategy. They combine margin-based sentence-level similarity to avoid semantic noise with word-embedding-based token weighting to refine the influence of demonstrations. 

Taken together, these studies show that ICL can improve machine translation under favourable conditions, but they also highlight its sensitivity to demonstration quality and distributional coverage. Importantly, most of this work evaluates languages with comparatively rich digital resources, leaving open the question of how reliably ICL-based MT behaviour transfers to truly low-resource languages with sparse data and unstable tokenization.

Recent work has explored structured linguistic scaffolding as a complement to standard few-shot prompting. \citet{lu2024chain} propose Chain-of-Dictionary Prompting (COD), which augments prompts with chained multilingual dictionary hints and reports large gains on FLORES-200. While effective, COD relies on proprietary models and dictionary resources that may not exist for many low-resource languages. In contrast, our approach uses open 7B-8B models and small parallel corpora, providing pivot translations as broader contextual scaffolding rather than word-level lexical hints.

Other work addresses low-resource adaptation through training-time methods. \citet{yong-etal-2023-bloom} show that adapter-based finetuning can outperform continued pretraining when adding new languages, with gains driven primarily by data availability. \citet{muennighoff2023crosslingual} introduce multitask prompted finetuning for multilingual models and demonstrate improved zero-shot generalization when prompt language aligns with the target. \citet{longpre2025atlas} analyze multilingual scaling laws and argue that at very low data scales, neither pretraining nor finetuning is computationally efficient. These approaches require supervised data and training compute, whereas our work targets inference-time prompting without parameter updates.

\subsection{Pivot languages aided LLM translation} 

Pivot strategies introduce an intermediate language to support translation in low-resource settings. Prior work has demonstrated that the choice of a pivot language can have significant impact on the translation quality.

Work by \citet{imamura-etal-2023-pivot} shows the poor zero-shot performance of multilingual NMT models translation can be enhanced by a using pivot language. In this work, they compare the pivot and direct translation using English as the pivot language. Their study also investigates which kind of parallel corpora is most effective to enhance multilingual pivot translation.

\citet{jiao2023chatgptgoodtranslatoryes} also evaluate ChatGPT for machine translation and introduce a pivot prompting strategy, in which the model first translates a source sentence into a high-resource pivot language before translating into the target language. They find that pivot prompting noticeably improves translation quality for distant or low-resource languages, and with GPT-4, ChatGPT achieves performance comparable to commercial translation systems even on some of the challenging language pairs.

Extending these ideas, \citet{ahmed-buys-2024-neural} introduces synthetic pivoting, where pivot sentences are generated from both the target and the source languages using the sequence level knowledge distillation. This approach reduces pivot translation complexity and improves BLEU scores for low-resource Southern African languages by up to 5.6 points.

Recent work by \citet{talwar2025pivotlanguagelowresourcemachine} highlight this in their study on Nepali-English translation, where Hindi is chosen as a pivot language due to its linguistic proximity to Nepali and the greater availability of Hindi parallel corpora. By employing both fully supervised transfer learning and semi-supervised back-translation, they show that using Hindi as a pivot language improves the Nepali-English translation baselines, emphasizing how a chosen pivot language can compensate for limited data availability.

\citet{lim2025mufu} reformulated low-resource translation as a post-editing task, where a teacher model generates auxiliary translations and a student model is finetuned to correct them, achieving strong gains on FLORES-200/NTREX. Their results suggest that even imperfect auxiliary translations can provide useful scaffolding. While Mufu relies on supervised finetuning, our work adapts this post-editing insight to pure ICL by using a single pivot translation combined with retrieved few-shot examples, without parameter updates or multi-model pipelines.

Collectively, these findings motivate our investigation into pivot language strategies for LLM translation. Our work builds on these insights by examining whether integrating pivot language examples and leveraging few-shot ICL can further enhance translation performance for languages like Konkani and Tunisian Arabic. In doing so, we aim to clarify the mechanisms by which pivot languages facilitate knowledge transfer in LLMs, while also extending the adaptation capabilities of models to new languages. This helps clarify when such approaches may, or may not, be effective for low-resource languages.

\section{Methodology}

We explore an inference-time technique of translation in settings where data, compute, and model scale are limited. Our goal is  to examine what kinds of evidence (such as linguistically related pivot languages and semantically retrieved few-shot examples) can be leveraged to support translation in an ICL setting using small (≈8B) decoder-only models, without fine-tuning or large parallel corpora. In particular, we investigate whether these signals provide useful guidance when translating into previously unseen or under-represented languages, and under what conditions they help, fail, or produce inconsistent behavior.

To support semantic retrieval, we construct a datastore of parallel translations organized as triplets consisting of an English source sentence, its pivot-language translation, and the corresponding target-language translation. These triplets are derived exclusively from the training split. We index the datastore using the English source sentences, as English is the input language at inference time. Sentence embeddings are computed using the \textbf{all-MiniLM-L12-v2} sentence transformer, which maps text into a dense vector space suitable for semantic similarity search. This representation allows semantically related translation examples to be clustered and retrieved efficiently.

\begin{figure}
    \centering
    \includegraphics[width=1.0\linewidth]{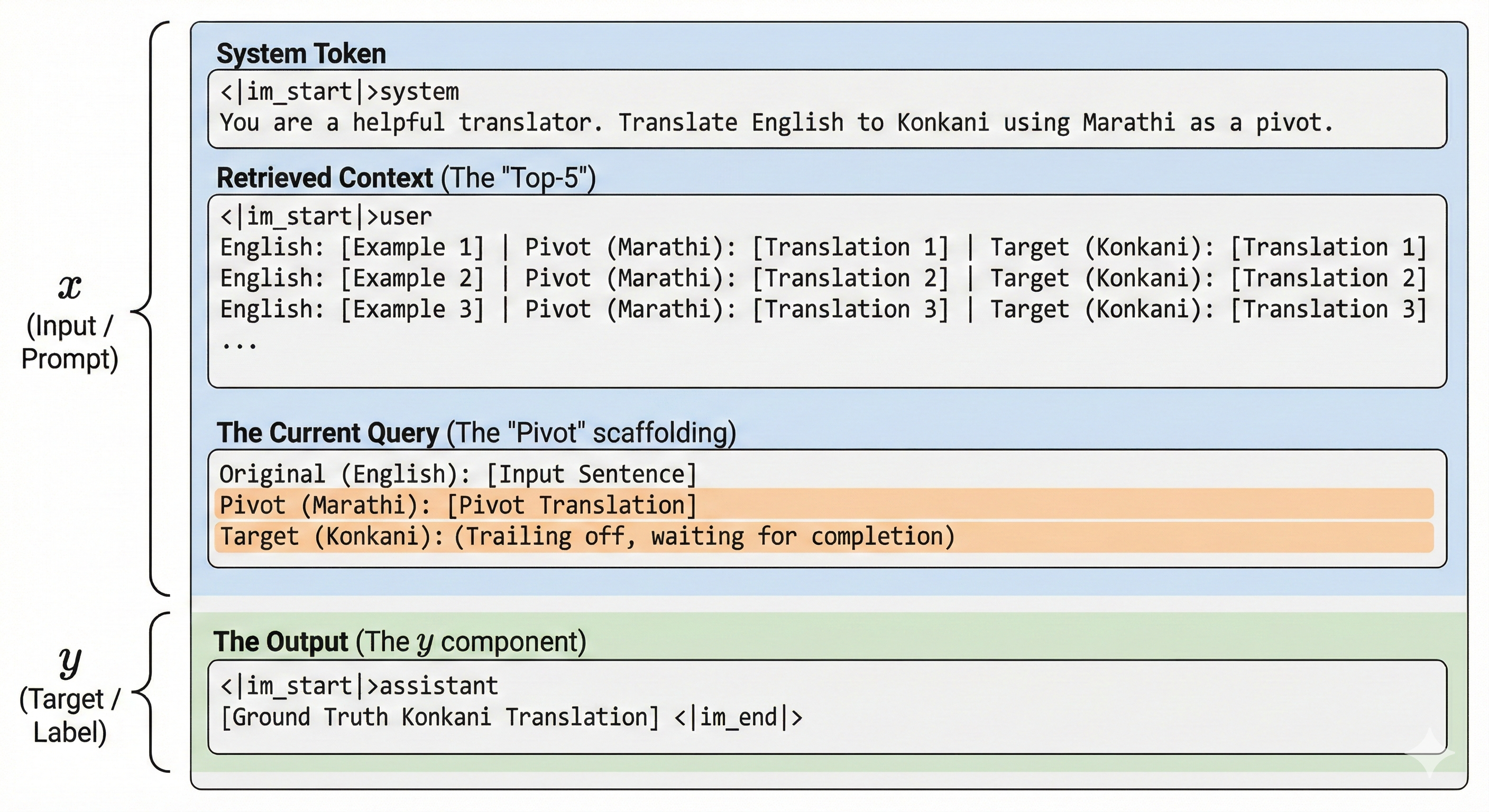}
    \caption{The Training Data Structure. The input $x$ contains the system instruction, retrieved semantic context (top-5), and the pivot scaffolding. The target $y$ contains only the model's generated translation.}
    \label{fig:placeholder}
\vspace{-2mm}
\end{figure}

At inference time, we generate an embedding for each input source sentence and query the vector datastore using cosine similarity. The top-$k$ most semantically similar triplets are retrieved and used as in-context demonstrations. These demonstrations are formatted as English-pivot-target examples and combined with the pivot translation corresponding to the same input sentence from the parallel corpus; no pivot translations are generated by the model or obtained from external sources. The resulting prompt is structured in ChatML format (see Appendix~\ref{appendix:prompt_template}). To mitigate contamination and retrieval leakage, all retrieved examples are drawn strictly from the training datastore, while evaluation is performed on a held-out test set that is never indexed or queried during retrieval. No test sentences or paraphrases are included in the datastore. We treat the number of in-context examples $k$ as a controllable parameter and evaluate its effect through a systematic ablation study (see Appendix~\ref{appendix:pivot-ablation}). This retrieval-augmented prompting workflow is illustrated in Figure~\ref{fig:placeholder}.

To isolate the contribution of the pivot language itself, we conduct controlled comparisons between pivot-augmented prompts and prompts constructed using identical retrieval procedures but excluding the pivot translation. This allows us to disentangle gains arising from semantic retrieval alone from those attributable to the linguistic bridge provided by the pivot language. The ablation study results without the pivot language are shown in
Table~\ref{tab:ablation_k_kn_direct} and Table~\ref{tab:ablation_k_tn_direct}.

% Overall, our methodology is designed to evaluate whether linguistically related pivot languages can facilitate inference-time knowledge transfer in LLMs under strict data and compute constraints, without parameter updates or reliance on large-scale parallel corpora.

\section{Experiments}

\subsection{Languages and data}

For this experiment, we focused on two low-resource languages. The first is Konkani, an Indian language spoken in the western part of India, with approximately 2.35 million speakers as of 2011. The second is Tunisian Arabic, spoken in Tunisia, with around 12 million speakers as of 2021. A motivation for selecting these languages was their use of non-Latin scripts. In addition, Tunisian Arabic is unique in the sense that, unlike regular Latin script which reads from left to right, this is from right to left. 

To provide a quantitative sanity check on pivot selection, we also compute
word-level Jaccard similarity between the pivot and target languages in our
datasets. This allows us to verify that the chosen pivots are lexically closer
to the target languages than English. A detailed description and full results
are provided in Appendix~\ref{appendix:jaccard}.

\subsection{Models}
In this experiment, we evaluate the performance of Unbabel's TowerInstruct-7B-v0.1 ~\citep{alves2024tower} and NousResearch's Hermes-2-Pro-Llama-3-8B ~\citep{teknium2024hermes}. The Hermes-2-Pro-Llama-3 model is a instruction-tuned version of Llama 3 ~\citep{llama3modelcard} known for its multilingual capabilities. Llama 3 supports 8 languages: English, German (deu), French (fra), Italian (ita), Portuguese (por), Hindi (hin), Spanish (spa), and Thai (tha), although the underlying foundation model has been trained on over 176 languages. Our aim is to see whether we can make use of the latent knowledge alignment of the model while translating to low resource languages. 

The TowerInstruct-7B-v0.1 model is finetuned from 
TowerBase-7B-v0.1 model. The 
TowerBase-7B-v0.1 model is continuously pretrained from from the Llama 2 model with a mixture of monolingual and parallel data with 20B tokens. The 
TowerBase-7B-v0.1 model is the finetuned with instruction dataset that is relevant to translation process. Some of these instruction tasks include Automatic Post Edition, Context-aware Translation, Named-entity Recognition etc. The languages supported by the 
TowerBase-7B-v0.1 model are English (eng), German (deu), French (fra), Dutch (nld), Italian (ita), Spanish (spa), Portuguese (por), Korean (kor), Russian (rus), and Chinese (zho). Although TowerInstruct-7B-v0.1 performs well on translation tasks, it is not expected to excel in languages it was not exposed to during training.

Our strategic selection of these models is designed to assess their effectiveness in translating languages outside their initial training sets. Despite TowerInstruct-7B-v0.1’s specialized training in translation, it has not been directly exposed to the specific low-resource languages focused on in this experiment, offering a unique test of its adaptability to unseen languages.

\subsection{Datasets}
% The Konkani parallel corpus was constructed using a dataset open-sourced by AI4Bharat, which also contributed to the training set for the IndicTrans2 model ~\citep{gala2023indictrans2}. This corpus includes English, Marathi, and Konkani.

% Given the experiment’s reliance on a pivot language included in the ICL examples, Marathi was selected as the pivot language due to its wider prevalence in western India and its linguistic similarity to Konkani. The test set for Konkani consisted of 200 records.

% For Tunisian Arabic, the corpus was derived from the work described in \citet{bouamor2014multidialectal} , with Modern Standard Arabic chosen as the pivot language. The parallel corpus for Tunisian Arabic contained 1,000 records, with 900 used in the training set and 100 used in the test set.

% Given that low-resource languages typically have limited available data, we conducted our study using a small training set of approximately 1,000 records to identify factors enhancing translation quality under stringent resource constraints.
We utilized two distinct multiparallel datasets, effectively organizing the data into aligned triplets (Source-Pivot-Target) to support our retrieval-augmented pipeline.

Konkani: We constructed a dataset of English-Marathi-Konkani triplets using the open-source corpus from AI4Bharat ~\citep{gala2023indictrans2}. Marathi was selected as the pivot language due to its linguistic similarity to Konkani and wider prevalence in western India. We created a distinct split of [800] examples for the training (retrieval) datastore and 200 examples for the held-out test set.

Tunisian Arabic: We derived a similar corpus of English-MSA-Tunisian triplets from the work described by \citet{bouamor2014multidialectal}, with Modern Standard Arabic (MSA) chosen as the pivot language. This dataset consists of 900 examples for the training datastore and 100 examples for the held-out test set.

In total, our study operates on small training sets of approximately 1,000 records per language. This constraint was chosen specifically to simulate realistic low-resource scenarios where large-scale parallel data is unavailable.

\section{Results}

\subsection{Does Pivot-Based Prompting Improve Translation?}

To establish a reference point, we first evaluate a direct prompting baseline, where the model is given only the English source sentence and instructed to translate directly into the target language, without access to a pivot language. In this setting, chrF++ scores are often extremely low (in some cases close to 1) because the models do not reliably generate text in the intended target language or script. Instead, the output frequently drifts toward better-represented neighboring languages (e.g., producing Marathi- or Hindi-like text when the target is Konkani, or MSA-like text for Tunisian Arabic). This behavior is observed across both Hermes and Tower, indicating that, without grounding signals, the model does not consistently infer the correct output language from the instruction alone.

We then compare this to our pivot-augmented prompting condition, in which the same input is supplied along with a translation into a linguistically related pivot language. In this setting, the few-shot demonstrations and pivot translation act as grounding signals that stabilize generation toward the intended script and language family. Tables~\ref{tab:konkani-results} and~\ref{tab:tunisian-results} report BLEU and chrF++ scores across three conditions (zero-shot ($k{=}0$), direct few-shot prompting without a pivot, and pivot-augmented prompting). For each configuration, we report the best-performing number of in-context examples ($k$), as determined in ablations in Appendix~\ref{tab:ablation_k_kn} and \ref{tab:ablation_k_tn}. 

For Konkani, introducing few-shot demonstrations, even without a pivot, leads to a substantial improvement in both chrF++ and BLEU, indicating that the examples themselves provide a strong anchoring effect for this previously unseen language. Adding the pivot language on top of these examples results in only small or mixed additional gains: for Hermes, the pivot condition yields a modest improvement over direct few-shot prompting (29.62→30.34 chrF++, 7.35→7.77 BLEU), whereas for Tower the pivot improves BLEU from 3.67 to 5.68, but does not improve chrF++. This suggests that, in this setting, most of the benefit arises from example-driven stabilization rather than from the pivot language itself.

For Tunisian Arabic, zero-shot scores are already relatively high, and both chrF++ and BLEU change marginally across the direct and pivot conditions, with no consistent advantage for either model. Here, few-shot prompting provides limited additional benefit, and the pivot language does not substantially alter model behavior, consistent with the interpretation that Tunisian Arabic which is already better represented in the underlying pretrained models.

We additionally evaluate whether using a pivot language that is explicitly supported by the model leads to improved translation quality. Given the constraints of our setup, the only configuration that satisfies this condition is Hindi as a pivot for Konkani using the Hermes-2-Pro-Llama-3-8B model. We analyze this setting in detail in Appendix~\ref{sec:llm_supported_pivot}, including token-to-word ratios and Jaccard similarity between Hindi and Konkani.

Across these experiments, we find that using a model-supported pivot language does not yield systematic improvements over linguistically motivated pivots such as Marathi. In several cases, performance degrades as the number of in-context examples increases, suggesting that native model support alone is insufficient to improve or stabilize low-resource translation.

To ensure that pivot-augmented prompting does not simply cause the model to reproduce pivot-language translations, we measure chrF overlap between the pivot outputs and the final generated translations. This analysis, reported in Appendix~\ref{sec:pivot_deviation}, shows consistently low chrF scores for both Konkani and Tunisian Arabic, indicating limited surface-level overlap between pivot and generated outputs. These results suggest that the model does not merely copy or lightly edit the pivot translation, but instead produces outputs that are substantially distinct from the pivot language.

One possible explanation, which we treat as hypothesis-generating rather than conclusive, comes from the token-to-word analysis in Table~\ref{tab:token_fertility}. For Tunisian Arabic, both models exhibit substantially lower token-to-word ratios (e.g., 4.96 vs.\ 7.65 for Tower; 2.16 vs.\ 4.09 for Hermes, comparing Aeb vs.\ Gom), indicating that the models segment Tunisian Arabic into fewer subword units than Konkani. Because Modern Standard Arabic (MSA) is well represented in most pretrained corpora, Tunisian Arabic, which shares script and lexical characteristics with MSA, may benefit indirectly from this representation. This would help explain why few-shot prompting and pivot augmentation yield smaller or inconsistent gains in this setting.

In contrast, the much higher token-to-word ratios for Konkani suggest a weaker lexical footprint in the pretrained vocabulary. Here, the few-shot examples and the pivot appear to act less as a source of additional translation competence and more as basic scaffolding for language identification, script adherence, and output stability.

However, we emphasize that this relationship is correlational rather than causal: tokenization efficiency alone does not fully explain performance differences, and other factors may contribute to the observed behavior.

\begin{table*}[t]
\centering

\begin{minipage}{0.48\textwidth}
\centering
\small
\begin{tabular}{l l c c}
\hline
\textbf{Model} & \textbf{Setting} & \textbf{BLEU} & \textbf{chrF++} \\
\hline
\textit{Baseline} & NLLB-200 & 7.51 & 26.82 \\
\hline
Hermes-2-Pro 
 & Zero-shot ($k{=}0$) & 1.49 & 1.30 \\
 & Direct (Best $k$) & 7.35 & 29.62 \\
 & \textbf{With Pivot (Best $k$)} & \textbf{7.77} & \textbf{30.34} \\
\hline
TowerInstruct 
 & Zero-shot ($k{=}0$) & 1.28 & 0.69 \\
 & Direct (Best $k$) & 3.67 & \textbf{21.25} \\
 & With Pivot (Best $k$) & \textbf{5.68} & 17.66 \\
\hline
\end{tabular}
\caption{Select Konkani translation results (Eng→Gom)}
\label{tab:konkani-results}
\end{minipage}
\hfill
\begin{minipage}{0.48\textwidth}
\centering
\small
\begin{tabular}{l l c c}
\hline
\textbf{Model} & \textbf{Setting} & \textbf{BLEU} & \textbf{chrF++} \\
\hline
\textit{Baseline} & NLLB-200 & 4.20 & 10.42 \\
\hline
Hermes-2-Pro 
 & Zero-shot ($k{=}0$) & 4.62 & 24.32 \\
 & \textbf{Direct (Best $k$)} & \textbf{6.27} & \textbf{24.32} \\
 & With Pivot (Best $k$) & 5.06 & 24.31 \\
\hline
TowerInstruct 
 & Zero-shot ($k{=}0$) & 4.19 & 17.62 \\
 & Direct (Best $k$) & 4.46 & \textbf{20.74} \\
 & With Pivot (Best $k$) & \textbf{4.99} & 20.63 \\
\hline
\end{tabular}
\caption{Select Tunisian Arabic results (Eng→Aeb)}
\label{tab:tunisian-results}
\end{minipage}

\vspace{-4mm}
\end{table*}

\subsection{Comparison to NLLB Reference Baselines}
As a point of external reference, we compare the best-performing few-shot and pivot-augmented scores in Tables~\ref{tab:konkani-results} and~\ref{tab:tunisian-results} with the NLLB-200 distilled baselines in Table~\ref{tab:nllb_baselines}. For Konkani, NLLB does not provide native support, and the baseline remains relatively low (26.82 chrF++, 7.51 BLEU). Our best Hermes pivot-augmented configuration attains 30.34 chrF++ and 7.77 BLEU, while the Tower pivot setting reaches 17.66 chrF++ and 5.68 BLEU. Thus, Hermes slightly exceeds the NLLB baseline on both metrics, whereas Tower remains below it.

For Tunisian Arabic, NLLB does include explicit support, but the baseline scores remain modest (10.42 chrF++, 4.20 BLEU). In contrast, both decoder-only LLMs achieve substantially higher performance even without fine-tuning: Hermes reaches 24.32 chrF++ and 6.27 BLEU (direct few-shot), and Tower reaches 20.63 chrF++ and 4.99 BLEU (pivot). This indicates that, even relative to a supervised MT system trained with explicit support for the language, few-shot prompting can yield stronger performance in this setting.

This contrast highlights a practical trade-off: improving NLLB performance for unsupported or weakly supported languages would typically require collecting supervised training data and fine-tuning the model, whereas our approach obtains measurable gains using only few-shot prompting with no parameter updates.

\subsection{Effect of Increasing the Number of In-Context Examples}
We next examine whether translation quality improves simply by increasing the number of retrieved demonstrations ($k$), independent of the pivot signal. For each model-language pair, we evaluate chrF++ and BLEU across multiple values of $k$ in both the direct and pivot-based translation settings (full results in Appendix~\ref{tab:ablation_k_kn}-\ref{tab:ablation_k_tn} and \ref{tab:ablation_k_kn_direct}-\ref{tab:ablation_k_tn_direct}).

For Konkani, increasing $k$ does not produce monotonic gains in either metric. In the direct translation setting, Hermes reaches its strongest chrF++ at $k{=}3$ (29.62) with relatively low BLEU (2.33), while performance drops at both smaller and larger $k$ values. Tower shows a similar pattern: chrF++ peaks at $k{=}2$ (21.25), while BLEU remains in the 3-4 range and collapses to 0 beyond $k{=}3$. In the pivot-based setting, Hermes attains its best chrF++ at $k{=}1$ (30.34) and best BLEU at $k{=}4$ (7.77), whereas Tower peaks at $k{=}2$ in BLEU (5.68) but achieves higher chrF++ at $k{=}3$ (17.66). Thus, for Konkani, both metrics improve relative to $k{=}0$, but additional demonstrations beyond the best-performing $k$ generally degrade performance.

A similar trend appears in Tunisian Arabic, although the baseline ($k{=}0$) performance is much stronger. In the direct setting, Hermes achieves its best BLEU at $k{=}1$ (6.27), while chrF++ remains highest at $k{=}0$ (24.32) and declines as $k$ increases. Tower exhibits modest variation across $k$, with chrF++ peaking at $k{=}4$ (20.74) despite little corresponding change in BLEU. In the pivot condition, Hermes again shows small fluctuations around its $k{=}0$ baseline (24.31 chrF++), while Tower reaches its highest BLEU at $k{=}2$ (4.99) and highest chrF++ at $k{=}5$ (20.63), before declining at larger $k$.

Across both languages and models, these results indicate that: Performance improves substantially when moving from $k{=}0$ to small $k$, but gains do not scale with additional demonstrations. ChrF++ and BLEU often peak at different values of $k$.

We hypothesize that one contributing factor is the interaction between $k$ and model context capacity. TowerInstruct operates effectively within a ~4K-token window, and performance often declines once prompts approach this length, suggesting truncation or overwriting effects. Hermes supports a larger context window, yet its performance likewise plateaus or degrades beyond moderate $k$, implying that the limitation is not purely architectural but also behavioral: models may underutilize long-range prompt structure or overweight spurious correlations from loosely related examples.

Taken together, these findings suggest that the gains observed in our experiments are not simply an artifact of “more examples.” Instead, a small number of semantically aligned demonstrations appears to provide most of the benefit, while additional examples can introduce noise that reduces both BLEU and chrF++. In settings where pivot-based prompting yields improvements, these effects should therefore be interpreted as complementary to, rather than interchangeable with, the contribution of few-shot demonstrations themselves.

\section{Limitations}

Many machine learning breakthroughs are enabled by an abundance of computational resources. However, access to large-scale compute is not uniformly available, including to most authors of this work. This disparity becomes even more apparent when working with communities that speak low-resource languages. Within these constraints, we aimed to rigorously test our hypotheses about pivot-based translation using the resources available to us. Importantly, these constraints also reflect realistic deployment conditions for many low-resource language communities, where access to large-scale compute, extensive annotation, and proprietary models is limited.

The primary limitation of this work is that, while we build on prior research on pivot languages to investigate whether linguistically related languages provide any useful signal for inference-time translation under resource constraints, the performance gains we observe are modest and often inconsistent. Working within our computational budget, we evaluated open-weight models in the 7B parameter range. While larger models may yield stronger performance, our results indicate that pivot-augmented prompting can sometimes improve performance, but its effects are highly sensitive to language characteristics and example selection, suggesting that further study is needed before drawing strong conclusions.

Additionally, much of the existing research on multilinguality and machine translation relies on human evaluation, which was not feasible in our setting. Under these constraints, and with respect for the communities that speak these languages, we evaluate how language models adapt to previously unseen languages in low-resource conditions using automatic metrics. We report BLEU and chrF++ scores computed with SacreBLEU~\citep{post-2018-call} for reproducibility (see Appendix~\ref{appendix:sacrebleu-signatures} for scoring signatures). 

However, these metrics have known limitations in low-resource and morphologically rich settings. As illustrated in Appendix~\ref{subsec:zero-bleu-example}, we observe cases where the generated Konkani translation is linguistically plausible and semantically related to the reference, yet differs substantially in surface form, resulting in very low BLEU and chrF++ scores. This highlights the brittleness of n-gram–based metrics for evaluating low-resource translation quality and motivates the need for human evaluation by native speakers to better capture semantic adequacy, pragmatic meaning, and dialectal correctness.

Another limitation is that our methodology depends on the availability of a high-resource pivot language that is linguistically similar to the target language, which restricts its applicability to languages without closely related pivots. While the approach is data-efficient, it also assumes access to high-quality parallel corpora; translation quality may degrade when there is a domain mismatch between the retrieved examples and the input text.

Given the promising results observed under these constrained settings, natural extensions of this work include scaling experiments to larger open-source models, conducting human-in-the-loop evaluations with native speakers, and exploring additional language pairs to better characterize the conditions under which pivot-augmented prompting helps, fails, or produces negligible effects.

\bibliography{custom}

\appendix

\section{Appendix}
\label{sec:appendix}

\subsection{Statistical Significance}
\label{sec:statistical-significance}

We conduct paired bootstrap resampling~\citep{koehn-2004-statistical} ($n{=}10{,}000$, $p < 0.05$) to test whether pivot prompting significantly outperforms direct translation. As shown in Table~\ref{tab:stat-sig}, no comparison reaches significance, indicating that observed trends are suggestive rather than conclusive.

% \begin{table}[h]
% \centering
% \scriptsize
% \setlength{\tabcolsep}{3pt}
% \begin{tabular}{@{}llc rr rr@{}}
% \toprule
% & & & \multicolumn{2}{c}{\textbf{BLEU}} & \multicolumn{2}{c}{\textbf{chrF++}} \\
% \cmidrule(lr){4-5} \cmidrule(lr){6-7}
% \textbf{Lang.} & \textbf{Mod.} & $k$ & $\Delta$ & $p$ & $\Delta$ & $p$ \\
% \midrule
% Gom & Her & $0$ & $+0.12$ & $.08$ & $+0.22$ & $.20$ \\
%     & Her & $1$ & $-0.18$ & $.88$ & $-0.89$ & $1.0$ \\
%     & Her & $2$ & $-0.12$ & $.75$ & $-0.33$ & $.89$ \\
%     & Tow & $1$ & $-0.02$ & $.57$ & $-0.84$ & $1.0$ \\
%     & Tow & $2$ & $+0.07$ & $.23$ & $-1.11$ & $1.0$ \\
% \midrule
% Aeb & Her & $0$ & $+0.23$ & $.07$ & $-0.05$ & $.56$ \\
%     & Her & $1$ & $+0.38$ & $.10$ & $-0.37$ & $.71$ \\
%     & Her & $2$ & $-0.26$ & $.78$ & $-1.23$ & $.98$ \\
%     & Tow & $0$ & $-0.02$ & $.55$ & $+0.02$ & $.48$ \\
%     & Tow & $1$ & $-0.20$ & $.72$ & $+0.34$ & $.23$ \\
%     & Tow & $2$ & $+0.13$ & $.31$ & $+0.25$ & $.22$ \\
% \bottomrule
% \end{tabular}
% \caption{Paired bootstrap significance (pivot $-$ direct). No comparison reaches $p < 0.05$. Gom: Konkani ($n{=}205$), Aeb: Tunisian Arabic ($n{=}100$). Her: Hermes, Tow: Tower.}
% \label{tab:stat-sig}
% \vspace{-4mm}
% \end{table}

\begin{table}
\centering
\footnotesize
\setlength{\tabcolsep}{2.5pt}
\begin{tabular}{llc rr rr}
\hline
 &  &  & \multicolumn{2}{c}{\textbf{BLEU}} & \multicolumn{2}{c}{\textbf{chrF++}} \\
\cline{4-5} \cline{6-7}
\textbf{Lang.} & \textbf{Mod.} & $k$ & $\Delta$ & $p$ & $\Delta$ & $p$ \\
\hline
Gom & Her & $0$ & $+0.12$ & $.08$ & $+0.22$ & $.20$ \\
    & Her & $1$ & $-0.18$ & $.88$ & $-0.89$ & $1.0$ \\
    & Her & $2$ & $-0.12$ & $.75$ & $-0.33$ & $.89$ \\
    & Tow & $1$ & $-0.02$ & $.57$ & $-0.84$ & $1.0$ \\
    & Tow & $2$ & $+0.07$ & $.23$ & $-1.11$ & $1.0$ \\
\hline
Aeb & Her & $0$ & $+0.23$ & $.07$ & $-0.05$ & $.56$ \\
    & Her & $1$ & $+0.38$ & $.10$ & $-0.37$ & $.71$ \\
    & Her & $2$ & $-0.26$ & $.78$ & $-1.23$ & $.98$ \\
    & Tow & $0$ & $-0.02$ & $.55$ & $+0.02$ & $.48$ \\
    & Tow & $1$ & $-0.20$ & $.72$ & $+0.34$ & $.23$ \\
    & Tow & $2$ & $+0.13$ & $.31$ & $+0.25$ & $.22$ \\
\hline
\end{tabular}
\caption{Paired bootstrap significance (pivot $-$ direct). No comparison reaches $p < 0.05$. Gom: Konkani ($n{=}205$), Aeb: Tunisian Arabic ($n{=}100$). Her: Hermes, Tow: Tower.}
\label{tab:stat-sig}
\vspace{-3mm}
\end{table}

\subsection{SacreBLEU Signatures and Reproducibility}
\label{appendix:sacrebleu-signatures}

All BLEU and chrF++ scores were computed using SacreBLEU~\citep{post-2018-call}. Following their recommendations, we report scoring signatures below for full reproducibility. Statistical significance was assessed via paired bootstrap resampling~\citep{koehn-2004-statistical} ($n{=}10{,}000$, $p < 0.05$); see Section~\ref{sec:statistical-significance}.

\begin{table}[h]
\centering
\small
\begin{tabular}{ll}
\hline
\textbf{Metric} & \textbf{Signature} \\
\hline
BLEU & \texttt{nrefs:1|case:mixed|eff:no|} \\
     & \texttt{tok:13a|smooth:exp|version:2.5.1} \\
\hline
chrF++ & \texttt{nrefs:1|case:mixed|eff:yes|} \\
       & \texttt{nc:6|nw:2|space:no|version:2.5.1} \\
\hline
\end{tabular}
\caption{SacreBLEU scoring signatures.}
\label{tab:sacrebleu-sigs}
\end{table} 

\subsection{NLLB baselines}
\label{appendix:nllb}

Table \ref{tab:nllb_baselines} reports reference translation scores from the NLLB-200 distilled model. NLLB is an encoder-decoder neural machine translation system trained for supervised MT, whereas the models in our study (Hermes and Tower) are decoder-only LLMs used in a few-shot, in-context prompting setting with no task-specific parameter updates. Accordingly, these numbers are provided only as contextual reference points rather than as directly comparable baselines. We also note that NLLB does not natively support Konkani; the scores reported for this variety in Table \ref{tab:nllb_baselines} reflect zero-shot transfer behavior rather than a tuned dialect model.
\begin{table}[H]
\centering
\renewcommand{\arraystretch}{1.15}
\begin{tabular}{lcc}
\hline
\textbf{Language Pair} & \textbf{BLEU} & \textbf{chrF++} \\
\hline
Eng-Gom & 7.51 & 26.82 \\
Eng-Aeb & 4.20 & 10.42 \\
\hline
\end{tabular}
\caption{NLLB-200 distilled reference baseline results for our evaluation
datasets.}
\label{tab:nllb_baselines}
\end{table}

\subsection{Token analysis}
As shown in Table~\ref{tab:token_fertility}, Hermes consistently exhibits lower token fertility than Tower across all non-English languages, particularly for low-resource and dialectal varieties, indicating more efficient subword representations. 
\begin{table}[htbp]
\centering

\begin{tabular}{llcc}
\hline
\textbf{Dataset} & \textbf{Language} & \textbf{Tower} & \textbf{Hermes} \\
\hline
Gom & Eng & 1.59 & 1.34 \\
Gom & Mar & 7.73 & 4.08 \\
Gom & Gom & 7.65 & 4.09 \\
\hline
Aeb & Eng & 1.27 & 1.21 \\
Aeb & MSA & 4.74 & 2.12 \\
Aeb & Aeb & 4.96 & 2.16 \\
\hline
\end{tabular}
\caption{Tokens per word across languages for Tower and Hermes models. Lower values indicate more efficient tokenization.}
\label{tab:token_fertility}
\end{table}

\subsection{Deviation from Pivot Translations}
\label{sec:pivot_deviation}
To assess whether the model simply reproduces pivot-language translations or instead generates genuinely distinct target-language outputs, we compute chrF scores between pivot translations and the final generated outputs for different values of $k$. chrF is well suited for this analysis, as it measures character-level overlap and is sensitive to direct copying, while remaining robust to morphological variation.

Table~\ref{tab:pivot-chrf-deviation} shows consistently low chrF scores across models, languages, and values of $k$, indicating limited surface-level overlap between pivot translations and generated output. This suggests that the models are not merely copying or lightly editing the pivot translations but are instead producing substantially different outputs.

Notably, the Tower model exhibits particularly low chrF scores compared to Hermes for both Arabic and Konkani, with scores for Konkani remaining below 11 across all values of $k$. This behavior indicates an even stronger departure from the pivot translations, reinforcing the conclusion that the generated outputs are not simple transcriptions or reformulations of the pivot language.

The stability of chrF scores across different values of $k$ further suggests that this divergence is systematic rather than an artifact of sampling variability. Overall, these results provide evidence that the generation step does not collapse to reproducing pivot-language translations, but instead yields outputs that are meaningfully distinct from the pivot representations.

\begin{table}[t]
\centering
\small
\setlength{\tabcolsep}{4pt}  % default is 6pt
\begin{tabular}{llccccc}
\toprule
\textbf{Language} & \textbf{Model} & \textbf{k=1} & \textbf{k=2} & \textbf{k=3} & \textbf{k=4} & \textbf{k=5} \\
\midrule
\multirow{2}{*}{Arabic}
 & Hermes & 24.09 & 24.89 & 25.17 & 23.79 & 24.06 \\
 & Tower  & 13.08 & 12.70 & 11.85 & 12.06 & 11.68 \\
\midrule
\multirow{2}{*}{Konkani}
 & Hermes & 28.96 & 27.45 & 27.82 & 25.95 & 26.80 \\
 & Tower  & 10.78 & 8.71  & 9.69  & 9.16  & 8.24  \\
\bottomrule
\end{tabular}
\caption{chrF scores between pivot translations and generated translations for different values of $k$. Lower scores indicate greater divergence from the pivot output, suggesting that the model is not simply reproducing the pivot translations.}
\label{tab:pivot-chrf-deviation}
\end{table}

\subsection{Jaccard similarity}
\label{appendix:jaccard}
Pairwise lexical similarity between the languages in our corpus is reported in
Tables~\ref{tab:jaccard-konkani} and \ref{tab:jaccard-arabic}.
Marathi and Konkani exhibit substantially higher lexical overlap than English with
either language, while Tunisian Arabic shows moderate overlap with Modern Standard
Arabic (MSA). These values are not used as a selection criterion, but rather serve as
supporting evidence that our chosen pivots are linguistically closer to the target
languages than English.

For each pair of languages, we compute word-level Jaccard similarity by treating the
vocabulary extracted from each corpus as a set. The similarity is defined as the size
of the intersection divided by the size of the union, yielding a value between 0 and 1,
where higher scores indicate greater lexical overlap.

In our datasets, Marathi (mar) and Konkani (Gom) show moderate lexical similarity
(10.6\%), while Tunisian Arabic (Aeb) and Modern Standard Arabic (Msa) exhibit higher
overlap (16.5\%). Notably, the Arabic variants show greater lexical closeness than the
Indo-Aryan language pair, reflecting the stronger typological affinity among Arabic
dialects.

A detailed breakdown of vocabulary sizes and pairwise similarity scores is presented
in Tables~\ref{tab:jaccard-konkani} and \ref{tab:jaccard-arabic}. In future work, we
plan to explore whether lexical similarity correlates with translation performance.

\begin{table}[t]
\centering

\begin{tabular}{lc}
\hline
\textbf{Language Pair} & \textbf{Jaccard Similarity} \\
\hline
Eng-Mar   & 0.0002 \\
Eng-Gom   & 0.0121 \\
Mar-Gom   & 0.1054 \\
\hline
\end{tabular}
\caption{Word-level Jaccard similarity scores for the Konkani corpus. Marathi and Konkani show substantially higher lexical overlap than English with either language.}
\label{tab:jaccard-konkani}
\end{table}

\begin{table}[t]
\centering

\begin{tabular}{lc}
\hline
\textbf{Language Pair} & \textbf{Jaccard Similarity} \\
\hline
Eng-MSA       & 0.0010 \\
Eng-Aeb  & 0.0010 \\
MSA-Aeb      & 0.1646 \\
\hline
\end{tabular}
\caption{Word-level Jaccard similarity scores for the Arabic corpus. MSA and Tunisian Arabic show moderate lexical overlap, while both exhibit minimal overlap with English.}
\label{tab:jaccard-arabic}
\end{table}

\subsection{Ablation on the Number of In-Context Examples ($k$) in the Direct Translation Setting}
\label{appendix:direct-ablation}

For completeness, we report ablations on the number of in-context examples 
($k$) in the \textbf{direct English→Target} setting, i.e., without the use of 
a pivot language. These results complement the pivot-based experiments and allow us to isolate the marginal effect 
of the pivot signal from the effect of in-context demonstrations alone.

Tables~\ref{tab:ablation_k_kn_direct} and \ref{tab:ablation_k_tn_direct}
show results for Konkani and Tunisian Arabic respectively, using the same retrieval and prompting setup as in the pivot configuration, but omitting the pivot translation from the prompt.

\subsubsection{Zero-BLEU but Non-Zero chrF++ Cases}
\label{subsec:zero-bleu-example}

During our direct translation experiments, we observed instances where
BLEU = 0 despite non-zero chrF++, particularly for Konkani. This occurs
when the model output diverges lexically from the reference despite
partial topical or semantic alignment. A representative example is shown
below.

\medskip
\noindent\textbf{Ground Truth (Konkani):}\\[0.5em]
\foreignlanguage{hindi}{बटाटां भितर मसालो भरसून ते बेसनाच्या पिठयेंत बुडोवन तेलांत बरे तळ्ळे कांय महाराष्ट्रांतलो हो सुवादीक आनी फामाद पदार्थ तयार जाता.}\\[1em]

\noindent\textbf{Model Translation:}\\[0.5em]
\foreignlanguage{hindi}{आलूच्या मसालेत संकरात बेसन बत्तर साठी अच्छा उत्साहसाठी महाराष्ट्रातील इतर प्रचारीक औषधे}

\begin{table*}[t]
  \centering

  \begin{tabular}{lccccc}
    \textbf{Model} & \textbf{Source} & \textbf{Target} & $k$ & \textbf{BLEU} & \textbf{chrF++} \\
    \hline
    \multicolumn{6}{c}{\textbf{Ablation: Number of In-Context Examples ($k$)}} \\
    \hline
    % -------- TowerInstruct with Marathi pivot --------
    \multicolumn{6}{l}{\emph{Unbabel/TowerInstruct-v0.1}} \\
    \hline
    Unbabel/TowerInstruct-v0.1 & Eng & Gom & 0  & 1.28 & 0.69 \\
    Unbabel/TowerInstruct-v0.1 & Eng & Gom & 1  & 3.67 & 21.01 \\
    Unbabel/TowerInstruct-v0.1 & Eng & Gom & 2  & 3.38 & 21.25 \\
    Unbabel/TowerInstruct-v0.1 & Eng & Gom & 3  & 3.39 & 19.30 \\
    Unbabel/TowerInstruct-v0.1 & Eng & Gom & 4  & 0.0 & 19.78 \\
    Unbabel/TowerInstruct-v0.1 & Eng & Gom & 5  & 0.0 & 19.38 \\
    \hline
    % -------- Hermes-2 with Marathi pivot --------
    \multicolumn{6}{l}{\emph{NousResearch/Hermes-2-Pro-Llama-3-8B}} \\
    \hline
    NousResearch/Hermes-2-Pro-Llama-3-8B & Eng & Gom & 0  & 1.49 & 1.30 \\
    NousResearch/Hermes-2-Pro-Llama-3-8B & Eng & Gom & 1  & 2.70 & 23.68 \\
    NousResearch/Hermes-2-Pro-Llama-3-8B & Eng & Gom & 2  & 2.72 & 23.87 \\
    NousResearch/Hermes-2-Pro-Llama-3-8B & Eng & Gom & 3  & 2.33 & 29.62 \\
    NousResearch/Hermes-2-Pro-Llama-3-8B & Eng & Gom & 4  & 1.90 & 28.75 \\
    NousResearch/Hermes-2-Pro-Llama-3-8B & Eng & Gom & 5  & 7.35 & 25.78 \\
    \hline
  \end{tabular}
  \caption{Ablation on the number of in-context examples ($k$) for English$\rightarrow$Konkani direct translation.}
  \label{tab:ablation_k_kn_direct}
\end{table*}

% \begin{table*}[t]
%   \centering
%   \begin{tabular}{lccccc}
%     \textbf{Model} & \textbf{Source} & \textbf{Target} & $k$ & \textbf{BLEU} & \textbf{chrF++} \\
%     \hline
%     \multicolumn{6}{c}{\textbf{Ablation: Number of In-Context Examples ($k$)}} \\
%     \hline
%     % -------- TowerInstruct with MSA pivot --------
%     \multicolumn{6}{l}{\emph{Unbabel/TowerInstruct-v0.1}} \\
%     \hline
%     Unbabel/TowerInstruct-v0.1 & Eng & Aeb & 1  & 3.67 & 23.69 \\
%     Unbabel/TowerInstruct-v0.1 & Eng & Aeb & 2  & 5.06 & 15.84 \\
%     Unbabel/TowerInstruct-v0.1 & Eng & Aeb & 3  & 4.62 & 17.09 \\
%     Unbabel/TowerInstruct-v0.1 & Eng & Aeb & 4  & 4.11 & 19.90 \\
%     Unbabel/TowerInstruct-v0.1 & Eng & Aeb & 5  & 4.27 & 21.56 \\

%     \hline
%     % -------- Hermes-2 with MSA pivot --------
%     \multicolumn{6}{l}{\emph{NousResearch/Hermes-2-Pro-Llama-3-8B}} \\
%     \hline
%     NousResearch/Hermes-2-Pro-Llama-3-8B & Eng & Aeb & 1  & 5.60 & 25.18 \\
%     NousResearch/Hermes-2-Pro-Llama-3-8B & Eng & Aeb & 2  & 4.62 & 18.45 \\
%     NousResearch/Hermes-2-Pro-Llama-3-8B & Eng & Aeb & 3  & 5.52 & 23.67 \\
%     NousResearch/Hermes-2-Pro-Llama-3-8B & Eng & Aeb & 4  & 3.42 & 20.96 \\
%     NousResearch/Hermes-2-Pro-Llama-3-8B & Eng & Aeb & 5  & 3.46 & 17.40 \\
    
%     \hline
%   \end{tabular}
%   \caption{Ablation on the number of in-context examples ($k$) for English$\rightarrow$Tn direct translation.}
%   \label{tab:ablation_k_tn_direct}
% \end{table*}
% Latest Experiment Results 
\begin{table*}[t]
  \centering
  \begin{tabular}{lccccc}
    \textbf{Model} & \textbf{Source} & \textbf{Target} & $k$ & \textbf{BLEU} & \textbf{chrF++} \\
    \hline
    \multicolumn{6}{c}{\textbf{Ablation: Number of In-Context Examples ($k$)}} \\
    \hline
    % -------- TowerInstruct with MSA pivot --------
    \multicolumn{6}{l}{\emph{Unbabel/TowerInstruct-v0.1}} \\
    \hline
    Unbabel/TowerInstruct-v0.1 & Eng & Aeb & 0  & 4.19 & 17.62 \\
    Unbabel/TowerInstruct-v0.1 & Eng & Aeb & 1  & 4.46 & 19.49 \\
    Unbabel/TowerInstruct-v0.1 & Eng & Aeb & 2  & 4.46 & 16.23 \\
    Unbabel/TowerInstruct-v0.1 & Eng & Aeb & 3  & 4.07 & 15.59 \\
    Unbabel/TowerInstruct-v0.1 & Eng & Aeb & 4  & 4.37 & 20.74 \\
    Unbabel/TowerInstruct-v0.1 & Eng & Aeb & 5  & 4.37 & 18.61 \\

    \hline
    % -------- Hermes-2 with MSA pivot --------
    \multicolumn{6}{l}{\emph{NousResearch/Hermes-2-Pro-Llama-3-8B}} \\
    \hline
    NousResearch/Hermes-2-Pro-Llama-3-8B & Eng & Aeb & 0  & 4.62 & 24.32 \\
    NousResearch/Hermes-2-Pro-Llama-3-8B & Eng & Aeb & 1  & 6.27 & 23.96 \\
    NousResearch/Hermes-2-Pro-Llama-3-8B & Eng & Aeb & 2  & 5.06 & 20.35 \\
    NousResearch/Hermes-2-Pro-Llama-3-8B & Eng & Aeb & 3  & 5.93 & 20.84 \\
    NousResearch/Hermes-2-Pro-Llama-3-8B & Eng & Aeb & 4  & 6.27 & 20.99 \\
    NousResearch/Hermes-2-Pro-Llama-3-8B & Eng & Aeb & 5  & 5.52 & 20.60 \\
    
    \hline
  \end{tabular}
  \caption{Ablation on the number of in-context examples ($k$) for English$\rightarrow$Tn direct translation.}
  \label{tab:ablation_k_tn_direct}
\end{table*}
% Latest Experiment Results 

\subsection{Ablation on the Number of In-Context Examples ($k$) in the Pivot-Based Setting}
\label{appendix:pivot-ablation}

For completeness, we report ablations on the number of in-context examples
($k$) in the \textbf{pivot-based} setting, i.e., where the model is provided
with a linguistically related pivot translation alongside the retrieved
few-shot examples allow us to examine the marginal contribution of the pivot signal. The performance of the models with pivot for konkani is shown in Table~\ref{tab:ablation_k_kn}
and Tunisian arabic Table~\ref{tab:ablation_k_tn}

\begin{table*}[t]
  \centering
  \begin{tabular}{lcccccc}
    \textbf{Model} & \textbf{Source} & \textbf{Pivot} & \textbf{Target} & $k$ & \textbf{BLEU} & \textbf{chrF++} \\

    \hline

    \multicolumn{7}{c}{\textbf{Ablation: Number of In-Context Examples ($k$) using Marathi Pivot (No Fine-Tuning)}} \\

    \hline

    % -------- TowerInstruct with Marathi pivot --------

    \multicolumn{7}{l}{\emph{Unbabel/TowerInstruct-v0.1}} \\

    \hline

    Unbabel/TowerInstruct-v0.1 & English & Mar & Gom & 0  & 2.07 & 1.30 \\

    Unbabel/TowerInstruct-v0.1 & English & Mar & Gom & 1  & 2.58 & 16.03 \\

    Unbabel/TowerInstruct-v0.1 & English & Mar & Gom & 2  & 5.68 & 8.94 \\

    Unbabel/TowerInstruct-v0.1 & English & Mar & Gom & 3  & 4.11 & 17.66 \\

    Unbabel/TowerInstruct-v0.1 & English & Mar & Gom & 4  & 2.84 & 4.90 \\

    Unbabel/TowerInstruct-v0.1 & English & Mar & Gom & 5  & 2.84 & 6.11 \\

    \hline

    % -------- Hermes-2 with Marathi pivot --------

    \multicolumn{7}{l}{\emph{NousResearch/Hermes-2-Pro-Llama-3-8B}} \\

    \hline

    NousResearch/Hermes-2-Pro-Llama-3-8B & English & Mar & Gom & 0  & 2.35 & 24.9 \\
    
    NousResearch/Hermes-2-Pro-Llama-3-8B & English & Mar & Gom & 1  & 3.49 & 30.34 \\

    NousResearch/Hermes-2-Pro-Llama-3-8B & English & Mar & Gom & 2  & 2.36 & 27.59 \\

    NousResearch/Hermes-2-Pro-Llama-3-8B & English & Mar & Gom & 3  & 2.72 & 25.89 \\

    NousResearch/Hermes-2-Pro-Llama-3-8B & English & Mar & Gom & 4  & 7.77 & 27.53 \\

    NousResearch/Hermes-2-Pro-Llama-3-8B & English & Mar & Gom & 5  & 5.73 & 28.65 \\

    \hline

  \end{tabular}

  \caption{Ablation on the number of in-context examples ($k$) for English$\rightarrow$Marathi$\rightarrow$Konkani translation.}

  \label{tab:ablation_k_kn}

\end{table*}

\begin{table*}[t]

  \centering

  \begin{tabular}{lcccccc}

    \textbf{Model} & \textbf{Source} & \textbf{Pivot} & \textbf{Target} & $k$ & \textbf{BLEU} & \textbf{chrF++} \\

    \hline

    \multicolumn{7}{c}{\textbf{Ablation: Number of In-Context Examples ($k$) using MSA Pivot (No Fine-Tuning)}} \\

    \hline

    % -------- TowerInstruct with MSA pivot --------

    \multicolumn{7}{l}{\emph{Unbabel/TowerInstruct-v0.1}} \\

    \hline

    Unbabel/TowerInstruct-v0.1 & Eng & Msa & Aeb & 0  & 4.37 & 16.45 \\

    Unbabel/TowerInstruct-v0.1 & Eng & Msa & Aeb & 1  & 3.46 & 18.74 \\

    Unbabel/TowerInstruct-v0.1 & Eng & Msa & Aeb & 2  & 4.99 & 16.57 \\

    Unbabel/TowerInstruct-v0.1 & Eng & Msa & Aeb & 3  & 4.77 & 17.32 \\

    Unbabel/TowerInstruct-v0.1 & Eng & Msa & Aeb & 4  & 3.09 & 19.80 \\

    Unbabel/TowerInstruct-v0.1 & Eng & Msa & Aeb & 5  & 3.75 & 20.63 \\

    \hline

    % -------- Hermes-2 with MSA pivot --------

    \multicolumn{7}{l}{\emph{NousResearch/Hermes-2-Pro-Llama-3-8B}} \\

    \hline

    NousResearch/Hermes-2-Pro-Llama-3-8B & English & Msa & Aeb & 0  & 5.06 & 24.31 \\
    
    NousResearch/Hermes-2-Pro-Llama-3-8B & English & Msa & Aeb & 1  & 4.93 & 21.27 \\

    NousResearch/Hermes-2-Pro-Llama-3-8B & English & Msa & Aeb & 2  & 3.74 & 18.18 \\

    NousResearch/Hermes-2-Pro-Llama-3-8B & English & Msa & Aeb & 3  & 4.20 & 20.17 \\

    NousResearch/Hermes-2-Pro-Llama-3-8B & English & Msa & Aeb & 4  & 4.93 & 19.42 \\

    NousResearch/Hermes-2-Pro-Llama-3-8B & English & Msa & Aeb & 5  & 4.77 & 16.32 \\

    \hline

  \end{tabular}

  \caption{Ablation on the number of in-context examples ($k$) for English$\rightarrow$Msa$\rightarrow$Aeb translation }

  \label{tab:ablation_k_tn}

\end{table*}

\subsection{Ablations with LLM-Supported Pivot Languages}
\label{sec:llm_supported_pivot}
Our experimental design selects pivot languages based on two primary criteria: (1) linguistic similarity to the target low-resource language, and (2) higher expected digital presence relative to the target. While we attempt to quantify pivot relevance using Jaccard similarity, this metric only imperfectly captures linguistic suitability, leaving gaps in systematic pivot selection. As an additional analysis, we consider pivot languages that are explicitly supported by the model, in order to examine whether native model support leads to improved translation performance.

A key limitation of this approach is that most general-purpose LLMs support only a narrow subset of languages, which substantially restricts coverage for low-resource targets. This constraint is evident even in our experimental setup: neither model explicitly supports Tunisian Arabic or closely related Arabic varieties, and for Konkani, only Hindi is supported, and only by the Hermes-2-Pro-Llama-3-8B model. Consequently, we evaluate this supported-pivot configuration only for Konkani and only under the Hermes-2-Pro-Llama-3-8B setting.

The Jaccard similarity between Hindi and Konkani (0.090) is slightly lower than that between Marathi and Konkani (0.105). However, because Hindi is explicitly supported by the model, this difference is reflected in tokenization behavior: the token-to-word ratio for Hindi under Hermes is substantially lower (2.85) than for Marathi and Konkani (both approximately 7, refer to Table~\ref{tab:token_fertility}), consistent with stronger lexical coverage in the pretrained vocabulary.

We report BLEU and chrF++ scores for this configuration in Table~\ref{tab:ablation_k_hin_kn_delta}. When comparing chrF++ scores against the corresponding Marathi-pivot setting, we do not observe systematic improvements from using a model-supported pivot language. In several cases, performance degrades substantially as the number of in-context examples increases, suggesting that native model support alone is insufficient to guarantee stable or improved pivot-based translation in this low-resource setting.

\begin{table*}[t]
  \centering
   \small 
  \begin{tabular}{lccccccc}
    \textbf{Model} & \textbf{Source} & \textbf{Pivot} & \textbf{Target} & $k$ & \textbf{BLEU} & \textbf{chrF++} & $\Delta$\textbf{chrF++} \\
    \hline

    \multicolumn{8}{c}{\textbf{Ablation: Number of In-Context Examples ($k$) using Hindi Pivot (No Fine-Tuning)}} \\
    \hline

    \multicolumn{8}{l}{\emph{NousResearch/Hermes-2-Pro-Llama-3-8B}} \\
    \hline

    NousResearch/Hermes-2-Pro-Llama-3-8B & English & Hin & Gom & 0  & 2.86 & 25.39 & +0.49 \\
    NousResearch/Hermes-2-Pro-Llama-3-8B & English & Hin & Gom & 1  & 2.47 & 24.12 & -6.22 \\
    NousResearch/Hermes-2-Pro-Llama-3-8B & English & Hin & Gom & 2  & 2.47 & 23.96 & -3.63 \\
    NousResearch/Hermes-2-Pro-Llama-3-8B & English & Hin & Gom & 3  & 2.41 & 23.69 & -2.20 \\
    NousResearch/Hermes-2-Pro-Llama-3-8B & English & Hin & Gom & 4  & 0.04 & 3.09  & -24.44 \\
    NousResearch/Hermes-2-Pro-Llama-3-8B & English & Hin & Gom & 5  & 0.02 & 2.49  & -26.16 \\
    \hline

  \end{tabular}
  \caption{Ablation on the number of in-context examples ($k$) for English$\rightarrow$Hindi$\rightarrow$Konkani translation using Nous Hermes. $\Delta$chrF++ is computed relative to the Marathi-pivot setting at the same $k$. Scores computed with SacreBLEU~\citep{post-2018-call}; signatures in Appendix~\ref{appendix:sacrebleu-signatures}.}
  \label{tab:ablation_k_hin_kn_delta}
\end{table*}

\subsection{Fine-Tuning Impact}

Our fine-tuning experiments were limited in scope and not comprehensive. Fine-tuning was performed on the same small training sets (~900 samples) used for few-shot example retrieval, without extensive hyperparameter tuning or architectural variations. While results show promise for Konkani, comprehensive fine-tuning ablations including varied training set sizes, learning rates, and LoRA configurations remain as future work.

\paragraph{Konkani:} In the finetuning experiments, we treat the zero-shot finetuned model (English→Konkani without pivot and without in-context demonstrations) as the reference baseline. For Hermes, the zero-shot finetuned condition achieves a chrF++ of 36.61, which increases to 40.17 when a Marathi pivot is introduced. For TowerInstruct, chrF++ increases from 17.39 (zero-shot finetuned without pivot) to 31.91 with pivot. For completeness, we also report few-shot finetuned results in Table~\ref{tab:finetune-gom}; across both settings, we observe consistent gains when the pivot language is incorporated during prompting.

 \paragraph{Tunisian Arabic:} As in the Konkani setting, we interpret the zero-shot finetuned model without a pivot as the reference baseline. For Hermes, the zero-shot finetuned condition achieves a chrF++ of 18.07, which increases to 21.87 when an MSA pivot is included during prompting. For TowerInstruct, chrF++ improves from 14.83 (zero-shot finetuned without pivot) to 19.16 with pivot. Few-shot finetuned results are also reported in Table~\ref{tab:finetune-aeb};  We again observe gains when the pivot language is incorporated.

Below we describe the experiment setting in detail: 

\paragraph{Hyperparameters:} With limited data, finetuning methods like prompt tuning (where embeddings are adjusted) or LoRA (Low-Rank Adaptation) prove particularly effective ~\citep{zhang2024scaling}. With Parameter-Efficient finetuning (PEFT), even increasing the data yielded modest performance improvements. For instance, using LoRA on the Hermes-2-Pro-Llama-3-8B model brought the trainable parameters down to 176,242,688, or just 2\% of the model’s total parameters. 

PEFT is computationally more efficient than pure ICL, which led us to adopt PEFT for our model fine-tuning process. We used the Huggingface Transformers library.

\begin{table}
\centering
\small
\begin{tabular}{l l}
\hline
\textbf{Parameter} & \textbf{Value} \\
\hline
\verb|batch_size|             & 1 \\
\verb|num_train_epochs|       & 1.5 \\
\verb|warmup_ratio|           & 0.03 \\
\verb|logging_steps|          & 25 \\
\verb|learning_rate|          & 2e-4 \\
\verb|gradient_checkpointing| & True \\
\verb|lr_scheduler_type|      & Cosine \\
\verb|weight_decay|           & 0.001 \\
\verb|save_strategy|          & No \\
\verb|optim|                  & \verb|PagedAdam| \\
\verb|warmup_steps|           & 100 \\
\verb|bf16|                   & True \\
\hline
\end{tabular}
\caption{Training parameters used in the model training process.}
\label{tab:training_parameters}
\vspace{-3mm}
\end{table}

% \begin{table}
% \begin{center}
% \begin{tabular}{|l|l|}
% \hline
% \bf Parameter & \bf Value \\
% \hline
% \verb|batch_size|  & 1 \\
% \verb|num_train_epochs|             & 1.5 \\
% \verb|warmup_ratio|                 & 0.03 \\
% \verb|logging_steps|               & 25 \\
% \verb|learning_rate|               & 2e-4 \\
% \verb|gradient_checkpointing|      & True \\
% \verb|lr_scheduler_type|           & Cosine \\
% \verb|weight_decay|                & 0.001 \\
% \verb|save_strategy|               & No \\
% \verb|optim|                       & \verb|PagedAdam| \\
% \verb|warmup_steps|                & 100 \\
% \verb|bf16|                        & True \\
% \hline
% \end{tabular}
% \end{center}
% \caption{\label{tab:training_parameters} Training parameters used in the model training process.}
% \end{table}

In addition, the model was loaded in 4-bit precision using the BitsAndBytes library with the nf4 quantization type. For fine-tuning, we employed the LoRA configuration, as detailed in the Table~\ref{tab:loraparameters}.

% \begin{table}
% \begin{center}
% \begin{tabular}{|l|p{4cm}|}
% \hline
% \bf Parameter & \bf Value \\
% \hline
% \texttt{r} & 64 \\
% \texttt{lora\_alpha} & 16 \\
% \texttt{lora\_dropout} & 0.1 \\
% \texttt{bias} & \texttt{none} \\
% \texttt{task\_type} & \texttt{CAUSAL\_LM} \\
% \texttt{target\_modules} & \texttt{['q', 'k', 'v', 'o', 'up', 'down', 'gate', 'lm\_head']} \\
% \hline
% \end{tabular}

% \end{center}
% \caption{\label{tab:loraparameters} LoRA configuration parameters.}
% \end{table}
\begin{table}
\centering
\small
\begin{tabular}{l p{4cm}}
\hline
\textbf{Parameter} & \textbf{Value} \\
\hline
\texttt{r} & 64 \\
\texttt{lora\_alpha} & 16 \\
\texttt{lora\_dropout} & 0.1 \\
\texttt{bias} & \texttt{none} \\
\texttt{task\_type} & \texttt{CAUSAL\_LM} \\
\texttt{target\_modules} & \texttt{['q', 'k', 'v', 'o', 'up', 'down', 'gate', 'lm\_head']} \\
\hline
\end{tabular}
\caption{LoRA configuration parameters.}
\label{tab:loraparameters}
\vspace{-3mm}
\end{table}

Parameters in Table~\ref{tab:inferenceparameters} were used to generate the output from the finetuned model during the evaluation.
\begin{table}
  \begin{center}
  \begin{tabular}{|l|} \hline
    \texttt{do\_sample}: \texttt{True} \\
    \texttt{temperature}: 0.1 \\
    \texttt{num\_return\_sequences}: 1 \\
    \texttt{max\_new\_tokens}: 200 \\
    \texttt{return\_full\_text}: \texttt{False} \\ \hline
  \end{tabular}
  \end{center}
  \caption{Inference parameters used for text generation.}
  \label{tab:inferenceparameters}
\end{table}
% Parameters in Table~\ref{tab:inferenceparameters} were used to generate the output
% from the fine-tuned model during evaluation.

% \begin{table}
% \centering
% \small
% \begin{tabular}{l}
% \hline
% \texttt{do\_sample}: \texttt{True} \\
% \texttt{temperature}: 0.1 \\
% \texttt{num\_return\_sequences}: 1 \\
% \texttt{max\_new\_tokens}: 200 \\
% \texttt{return\_full\_text}: \texttt{False} \\
% \hline
% \end{tabular}
% \caption{Inference parameters used for text generation.}
% \label{tab:inferenceparameters}
% \vspace{-3mm}
% \end{table}

\subsection{Prompt Template}

\label{appendix:prompt_template}

Both the TowerInstruct-7B-v0.1 model and Hermes-2-Pro-Llama-3-8B model utilize a similar prompt format. The full prompt format is below. 
\begin{tcolorbox}[breakable]
\begin{figure}[H]
\centering
\begin{verbatim}
<|im_start|>user
APE is a task designed to enhance
the quality of the translation 
by performing minor adjustments
Original (English): [Original text]
Translation: [Pivot language]
Post-edited:
<|im_end|>
<|im_start|>assistant
[LLM translation]
<|im_end|>
\end{verbatim}
\end{figure}
\end{tcolorbox}
The prompt includes the source sentence in English and its translation in a pivot language. For in-context learning, the prompt contains five demonstrations. In each demonstration, the assistant field is pre-filled with the target language translation. These demonstrations are carefully selected sentences that closely resemble the sentence to be translated. In the final instance, the assistant field is left blank. This prompt structure proved to be highly effective for translation tasks of this nature. However, when using this format with the base model, the outputs often included elements like “Note,” gibberish, and repetitions. After fine-tuning the model with this format, the generated translations adhered to the expected structure and consistently produced Konkani sentences.

\subsection{Translation APE Examples}

\begin{itemize}
\item \textbf{Tunisian Example Prompt:}
<|begin\_of\_text|><|im\_start|>user:
APE is a task designed to enhance the quality of the translation by performing only minor adjustments to fix any existing translation mistakes. If the translation is already correct, you should retain it as is.

\textbf{Original (English):} always and always \\[1em]
\textbf{Translation (Modern Standard Arabic):} \foreignlanguage{arabic}{دائماً} \\[1em]
\textbf{Post-edited (Tunisian):} <|im\_end|>

<|im\_start|>assistant:
\foreignlanguage{arabic}{ابدا ابدا} <|im\_end|>

<|im\_start|>user:
APE is a task designed to enhance the quality of the translation by performing only minor adjustments to fix any existing translation mistakes. If the translation is already correct, you should retain it as is.

\textbf{Original (English):} there a lot of things that tell us shut up and brake us... \\[1em]
\textbf{Translation (Modern Standard Arabic):} \foreignlanguage{arabic}{هناك الكثير من الاشياء التى تقول لنا ان نصمت و تعترض طريقنا} \\[1em]
\textbf{Post-edited (Tunisian):} <|im\_end|>

<|im\_start|>assistant:
\foreignlanguage{arabic}{فما برشا حاجات تسكتنا و توقفنا} <|im\_end|>

<|im\_start|>user:
APE is a task designed to enhance the quality of the translation by performing only minor adjustments to fix any existing translation mistakes. If the translation is already correct, you should retain it as is.

\textbf{Original (English):} And sometime no \\[1em]
\textbf{Translation (Modern Standard Arabic):} \foreignlanguage{arabic}{وقليلاً لا} \\[1em]
\textbf{Post-edited (Tunisian):} <|im\_end|>

<|im\_start|>assistant:
\foreignlanguage{arabic}{و ساعات لا} <|im\_end|>

<|im\_start|>user:
APE is a task designed to enhance the quality of the translation by performing only minor adjustments to fix any existing translation mistakes. If the translation is already correct, you should retain it as is.

\textbf{Original (English):} like I said before, in good and in bad \\[1em]
\textbf{Translation (Modern Standard Arabic):} \foreignlanguage{arabic}{كما قلت قبل ذلك هناك الجيد وهناك الشرير} \\[1em]
\textbf{Post-edited (Tunisian):} <|im\_end|>

<|im\_start|>assistant:
\foreignlanguage{arabic}{كيما قلت قبل في الحلو و الخايب} <|im\_end|>

<|im\_start|>user:
APE is a task designed to enhance the quality of the translation by performing only minor adjustments to fix any existing translation mistakes. If the translation is already correct, you should retain it as is.

\textbf{Original (English):} let us be really happy away from standard stuffs \\[1em]
\textbf{Translation (Modern Standard Arabic):} \foreignlanguage{arabic}{اتركنا نسعد حقاً بعيد عن التابوهات} \\[1em]
\textbf{Post-edited (Tunisian):} <|im\_end|>

<|im\_start|>assistant:
\foreignlanguage{arabic}{خلينا نفرح برسمي بعيد على كل شي} <|im\_end|>

<|im\_start|>user:
APE is a task designed to enhance the quality of the translation by performing only minor adjustments to fix any existing translation mistakes. If the translation is already correct, you should retain it as is.

\textbf{Original (English):} we shouldn't be negative all the time \\[1em]
\textbf{Translation (Modern Standard Arabic):} \foreignlanguage{arabic}{لا يجب ان نكون بهذه السلبيه على طول الدوام.} \\[1em]
\textbf{Post-edited (Tunisian):} <|im\_end|>

<|im\_start|>assistant:
Translation: <|im\_end|>

\textbf{Response from the model setting with the highest Chrf++ score:}\\[1em]
\foreignlanguage{arabic}{ما لازم نكون سلبيين على طول الوقت}

 \item \textbf{Konkani Example Prompt:}

<|begin\_of\_text|><|im\_start|>user:
APE is a task designed to enhance the quality\\ of the translation by performing\\ only minor adjustments to fix any existing translation mistakes. If the translation is already correct, you should retain it as is.

\textbf{Original (English):} Great was his compassion for the two dear ones at this parting moment. \\[1em]
\textbf{Translation (Marathi):} \foreignlanguage{hindi}{विलग होताना त्याच्या दोन प्रिय व्यक्तींविषयी त्याला अतीव करुणा वाटत होती.} \\[1em]
\textbf{Post-edited (Konkani):} <|im\_end|>

<|im\_start|>assistant:
\foreignlanguage{hindi}{जिवाभावाच्या दोगांयचो त्याग करपी त्या खिणावेळार ताची करुणा सुमराभायली आशिल्ली.} <|im\_end|>

<|im\_start|>user:
APE is a task designed to enhance the quality of the translation by performing only minor adjustments to fix any existing translation mistakes. If the translation is already correct, you should retain it as is.

\textbf{Original (English):} Suleman's parents were quite tall. \\[1em]
\textbf{Translation (Marathi):} \foreignlanguage{hindi}{सुलेमानचे पालक बरेच उंच होते.} \\[1em]
\textbf{Post-edited (Konkani):} <|im\_end|>

<|im\_start|>assistant:
\foreignlanguage{hindi}{सुलेमानाचे पालक खूब उंच आशिल्ले.} <|im\_end|>

<|im\_start|>user:
APE is a task designed to enhance the quality of the translation by performing only minor adjustments to fix any existing translation mistakes. If the translation is already correct, you should retain it as is.

\textbf{Original (English):} Our country owes a deep debt of gratitude to our valiant ex-Servicemen. \\[1em]
\textbf{Translation (Marathi):} \foreignlanguage{hindi}{आपल्या शूर माजी सैनिकांप्रति आपला देश कृतज्ञतेने अपार ऋणी आहे.} \\[1em]
\textbf{Post-edited (Konkani):} <|im\_end|>

<|im\_start|>assistant:
\foreignlanguage{hindi}{आमचो देश शूरवीर सेवानिवृत्त-सैनिकांक कृतज्ञतायेचें रीण देणें आसा.} <|im\_end|>

<|im\_start|>user:
APE is a task designed to enhance the quality of the translation by performing only minor adjustments to fix any existing translation mistakes. If the translation is already correct, you should retain it as is.

\textbf{Original (English):} Boys are equally vulnerable to sexual abuse. \\[1em]
\textbf{Translation (Marathi):} \foreignlanguage{hindi}{मुलगेही लैंगिक छळाला तेवढेच बळी पडू शकतात.} \\[1em]
\textbf{Post-edited (Konkani):} <|im\_end|>

<|im\_start|>assistant:
\foreignlanguage{hindi}{चलेय लैगिक अत्याचाराची तितलीच शिकार जावंक शकतात.} <|im\_end|>

<|im\_start|>user:
APE is a task designed to enhance the quality of the translation by performing only minor adjustments to fix any existing translation mistakes. If the translation is already correct, you should retain it as is.

\textbf{Original (English):} But Mangal Pandey's brave deed was done through devotion to a high and noble principle. \\[1em]
\textbf{Translation (Marathi):} \foreignlanguage{hindi}{पण मंगल पांडेची शौर्यशाली कृती ही एका उच्च आणि उदात्त तत्त्वाप्रतिच्या समर्पणातून केली गेली होती.} \\[1em]
\textbf{Post-edited (Konkani):} <|im\_end|>

<|im\_start|>assistant:
\foreignlanguage{hindi}{पूण मंगल पांडेचें धाडशी कर्तुब एके उंचेल्या आनी उदार तत्वनिश्ठेचें आशिल्लें.} <|im\_end|>

<|im\_start|>user:
APE is a task designed to enhance the quality of the translation by performing only minor adjustments to fix any existing translation mistakes. If the translation is already correct, you should retain it as is.

\textbf{Original (English):} The brothers were deeply attached to each other. \\[1em]
\textbf{Translation (Marathi):} \foreignlanguage{hindi}{भाऊ एकमेकांना खूप जवळ होते.} \\[1em]
\textbf{Post-edited (Konkani):} <|im\_end|>

<|im\_start|>assistant:
Translation: <|im\_end|>

%[2em]
\textbf{Response from the model setting with the highest Chrf++ score:}\\[1em]
\foreignlanguage{hindi}{भावांनी एकमेकांकडेन खूब नजीकाय आशिल्ली.}
\end{itemize}

\begin{table*}[t]
  \centering
  \begin{tabular}{lcccccc}
    \hline
    \textbf{Model} & \textbf{Source} & \textbf{Pivot} & \textbf{Target} & \textbf{BLEU} & \textbf{CHRF++} \\
    \hline
    \multicolumn{6}{c}{\textbf{Few-shot Finetuned}} \\
    \hline
    Unbabel/TowerInstruct-v0.1 & English & - & Konkani & 4.18 & 31.57 \\
    NousResearch/Hermes-2-Pro-Llama-3-8B & English & - & Konkani & 3.49 & 31.49 \\
    Unbabel/TowerInstruct-v0.1 & English  & Marathi & Konkani & 7.80 & 17.60 \\
    NousResearch/Hermes-2-Pro-Llama-3-8B & English & Marathi & Konkani & 12.14 & 34.92 \\
    \hline
    \multicolumn{6}{c}{\textbf{Zero-shot Finetuned}} \\
    \hline
    Unbabel/TowerInstruct-v0.1 & English & - & Konkani & 1.89 & 17.39 \\
    NousResearch/Hermes-2-Pro-Llama-3-8B & English & - & Konkani & 4.01 & 36.61 \\
    Unbabel/TowerInstruct-v0.1 & English & Marathi & Konkani & 7.94 & 31.91 \\
    NousResearch/Hermes-2-Pro-Llama-3-8B & English & Marathi & Konkani & 8.38 & 40.17 \\
    \hline
  \end{tabular}
  \caption{Performance comparison of finetuned models in few-shot and zero-shot settings for Konkani translation.}
  \label{tab:finetune-gom}
\end{table*}

\begin{table*}[t]
  \centering
  \begin{tabular}{lccccc} 
    \textbf{Model} & \textbf{Source} & \textbf{Pivot} & \textbf{Target} & \textbf{BLEU} & \textbf{CHRF++} \\
    \hline
    \multicolumn{6}{c}{\textbf{Few-shot Finetuned}} \\
    \hline
    Unbabel/TowerInstruct-v0.1 & English & - & Tn & 3.3 & 21.05 \\
    NousResearch/Hermes-2-Pro-Llama-3-8B & English & - & Tn & NA & NA \\
    Unbabel/TowerInstruct-v0.1 & English & Msa & Tn & 2.82 & 17.12 \\
    NousResearch/Hermes-2-Pro-Llama-3-8B & English & Msa & Tn & {8.02} & {35.99} \\
    \hline
    \multicolumn{6}{c}{\textbf{Zero-shot Finetuned}} \\
    \hline
    Unbabel/TowerInstruct-v0.1 & English & - & Tn & 1.48 & 14.83 \\
    NousResearch/Hermes-2-Pro-Llama-3-8B & English & - & Tn & 5.02 & 18.07 \\
    Unbabel/TowerInstruct-v0.1 & English & Msa & Tn & 2.09 & 19.16 \\
    NousResearch/Hermes-2-Pro-Llama-3-8B & English & Msa & Tn & 4.62 & {21.87} \\
    \hline
  \end{tabular}
  \caption{Performance comparison of finetuned models in few-shot and zero-shot settings for Tunisian Arabic translation.}
  \label{tab:finetune-aeb}
\end{table*}

\end{document}